\documentclass[10pt,twocolumn,letterpaper]{article}
\usepackage[letterpaper,textwidth=6.875in,textheight=8.875in,top=1in]{geometry}
\usepackage{times}
\usepackage{amsmath,amssymb,graphicx,booktabs}
\usepackage[numbers,sort&compress]{natbib}
\usepackage[font=small,labelsep=period]{caption}
\usepackage[hyphens]{url}
\usepackage[hidelinks]{hyperref}
\title{Depth-Dominant Skeleton Detection for Natural Scenes}

\author{Chengkun Rao \quad Xinrong Hu \quad Min Li \quad Yangjun Ou \quad Ziwei Luo \quad Ye Li\\
Zhaojing Wang \quad Junwei Tang \quad Bangchao Wang \quad Xiaoyun Yan\thanks{Corresponding author.} \\
School of Computer Science and Artificial Intelligence\\
Wuhan Textile University\\
{\tt\small 2304300206@wtu.edu.cn, \ xyyan@wtu.edu.cn}
}

\date{}

\begin{document}
\maketitle
\begin{abstract}
Existing natural-scene skeleton detectors follow an RGB-representation-based feature encoding paradigm: they locate skeletons within the RGB appearance space, in which the skeleton's shape and topology can only be inferred implicitly from entangled color and texture cues. Despite notable progress, this encoding paradigm degrades markedly on complex scenes such as cluttered or occluded ones, where appearance distractors overwhelm the weak structural signal. We observe that depth images are inherently insensitive to color and texture and provide clear regional contours and inter-region spatial relationships, which naturally eases the difficulty of skeleton detection in complex scenes. Motivated by this, we propose for the first time a novel skeleton detection paradigm built on depth-representation-dominant feature encoding, in which depth dominates and RGB serves as the auxiliary, and present a model, DDSkel, under this paradigm. DDSkel adopts an asymmetric encoder design that fuses the feature representation of RGB images into that of depth images, with the RGB branch containing only 23\% of the parameters of the depth branch. DDSkel has a simple structure without intricate designs. Nevertheless, with only about 19\% of the trainable parameters of the largest state-of-the-art methods, DDSkel attains the highest F-measure scores on four mainstream benchmarks, namely SK506, SK-LARGE, WH-SYMMAX, and SymPASCAL. Extensive experiments suggest that for natural-image skeleton detection, the depth-dominant paradigm is a more suitable choice than the RGB-based paradigm, achieving better performance with substantially fewer parameters. 
\end{abstract}
\section{Introduction}
\label{sec:intro}

Generally speaking, a classic skeleton, a key technique for object shape representation, is the medial axis of an object, reduced to a single-pixel-wide line while preserving its topological structure and key shape features \cite{Blum}. Early skeleton detection methods process binary images only \cite{Bai01, Sana01}; however, as its application scope expanded, especially in the era of deep learning, an increasing number of algorithms have begun to extract skeletons from natural images \cite{Shen01, Zhang01, Bai02, Wang03}. Skeleton detection in natural images has a wide range of applications, such as object recognition \cite{Li01}, text processing \cite{Xie01}, medical image analysis \cite{Chen01}, object detection \cite{Gupta01}, industrial inspection \cite{Guo01}, pose estimation \cite{Xu03}, and remote sensing analysis \cite{Zhou01}.

\begin{figure}[t]
  \centering
   \includegraphics[width=1\linewidth]{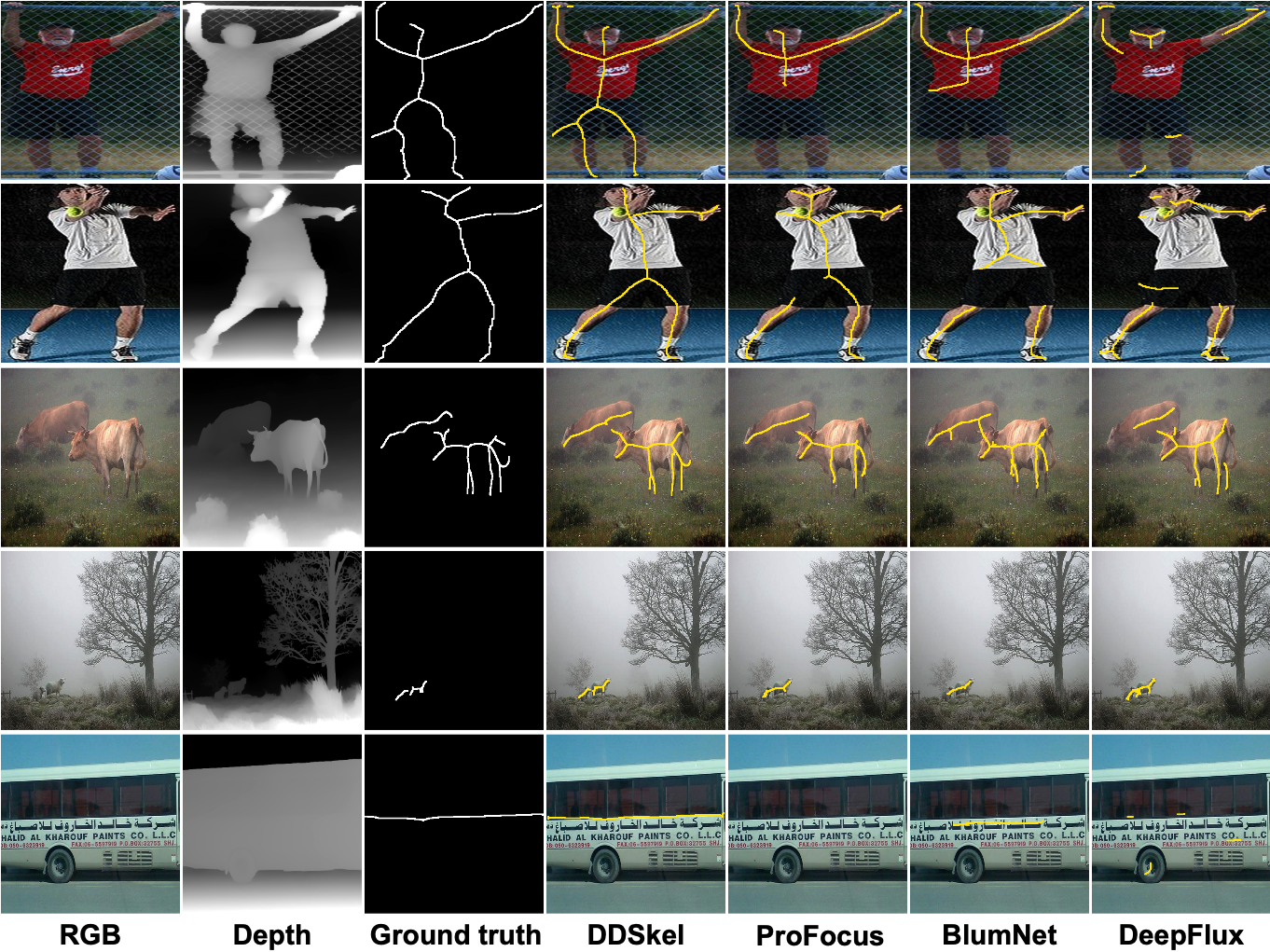}
   \caption{Comparison of our DDSkel with three state-of-the-art methods
(ProFocus~\cite{Wang03}, BlumNet~\cite{Zhang01}, and
DeepFlux~\cite{Wang01}); yellow lines denote detected skeletons.}
   \label{fig:FIG1}
\end{figure}

All existing natural-scene skeleton detectors follow an
RGB-representation-based feature encoding paradigm: they locate skeletons
directly within the RGB appearance space, so the shape and topology are
recovered implicitly from color and texture. Despite the remarkable progress,
methods under this paradigm still degrade markedly on cluttered or occluded
scenes. Fig.~\ref{fig:FIG1} illustrates this phenomenon, where the columns
show the test images, the depth images generated by the DepthAnythingV2-Small model \cite{Yang04}, the ground
truth, and the skeletons predicted by our proposed model, DDSkel, and three state-of-the-art
methods (ProFocus~\cite{Wang03}, BlumNet~\cite{Zhang01}, and
DeepFlux~\cite{Wang01}); the yellow lines denote the detected skeletons.
These challenging images contain foreground objects with background-like
color or texture, partial occlusion, or complex content, and none of the
competing methods yields satisfactory detections.

The phenomenon in Fig.~\ref{fig:FIG1} is rooted in the inherent nature of
skeleton detection: its goal is to extract the medial axis of object shapes,
which is independent of categories, textures, colors, and background
appearance. However, RGB images, especially in complex scenes, inevitably
introduce substantial shape-irrelevant interference: cluttered textures,
foreground-background appearance confusion, and occlusion-induced ambiguity,
which are the root cause of the drastic performance degradation in complex
scenarios. To mitigate these issues, existing methods have to design more
complex networks, yet the limited scale of current datasets increases the
risk of overfitting.

We notice that depth images are inherently insensitive to color and texture cues, and can provide region contours as well as spatial depth relationships across regions, which naturally alleviates the inherent difficulty of skeleton detection in complex scenes. The second column of Fig.\ref{fig:FIG1} clearly demonstrates three key advantages of depth images: (1) depth carries minimal redundant detail in foreground and background, strongly suppressing shape-irrelevant clutter; (2) it captures object contours, which are critical cues for skeleton detection, more reliably than RGB images; and (3) it encodes scene geometry, facilitating detection under occlusion and multiple overlapping objects.

In recent years, monocular depth estimation has matured rapidly \cite{Yang04}. State-of-the-art methods in this field are now capable of producing high-quality depth images, such as DepthAnythingV2-Small, which facilitates the direct use of depth images on existing skeleton detection datasets without requiring dedicated depth capture.

Nevertheless, depth images inherently lack category-level semantics, while skeleton ground truth is only annotated on objects of specific categories; RGB images are therefore still required to provide semantic guidance. Moreover, RGB images can serve as a compensatory fallback in regions where the depth images degrade. In other words, the RGB modality is not redundant, but it is auxiliary.

Based on the above analysis, this paper breaks away from the long-standing RGB-representation-based feature encoding paradigm and proposes a novel depth-representation-dominant feature encoding paradigm, in which a depth image dominates the encoding process, while an RGB image acts as the auxiliary. This paradigm fully leverages the complementary strengths of depth and RGB modalities for skeleton detection. Guided by this paradigm, we present DDSkel, a skeleton detection model that takes RGB images and corresponding depth images as input.

DDSkel adopts an encoder-decoder architecture. Its encoder employs an asymmetric dual-branch design, unidirectionally fusing RGB features into depth features through the Cross-stage Fusion Module (CFM) and the Dual-modality Attention-based Fusion Module (DAFM), with the auxiliary RGB branch containing only 23\% of the parameters of the dominant depth branch. The augmented depth features are then fed into the decoder for skeleton prediction. To demonstrate the advantages of the new paradigm, we keep DDSkel as simple as possible; its encoder and decoder contain approximately 19M and 0.35M trainable parameters, respectively. With only about 19\% of the trainable parameters of the largest state-of-the-art methods, DDSkel surpasses all existing methods in F-measure on four mainstream benchmarks, namely SK506 \cite{Shen02}, SK-LARGE \cite{Shen01}, WH-SYMMAX \cite{Shen03}, and SymPASCAL \cite{Ke01}.

The main contributions of this paper are as follows:
\begin{itemize}
\item We propose a novel depth-representation-dominant feature encoding paradigm to overcome the limitations of the long-standing RGB-representation-based feature encoding paradigm for skeleton detection in natural scenes.
\item We present DDSkel, a new skeleton detection model built under this paradigm, which takes RGB images and corresponding depth images as input. Its encoder is asymmetric, pairing a dominant depth backbone with an auxiliary RGB backbone, where RGB features are unidirectionally fused into the depth features for skeleton prediction by the decoder. DDSkel has a simple structure without intricate designs.
\item Extensive experiments show that, despite its simple structure, DDSkel
achieves the highest F-measure scores on all four mainstream benchmarks,
surpassing existing state-of-the-art approaches, with only about 19\% of
the trainable parameters of the largest state-of-the-art methods. The experiments also suggest that the new paradigm is a more suitable choice than the RGB-based paradigm for natural-image skeleton detection.
\end{itemize}

\section{Related Work}

\begin{figure*}[t]
	\centering
	\includegraphics[width=1.9\columnwidth]{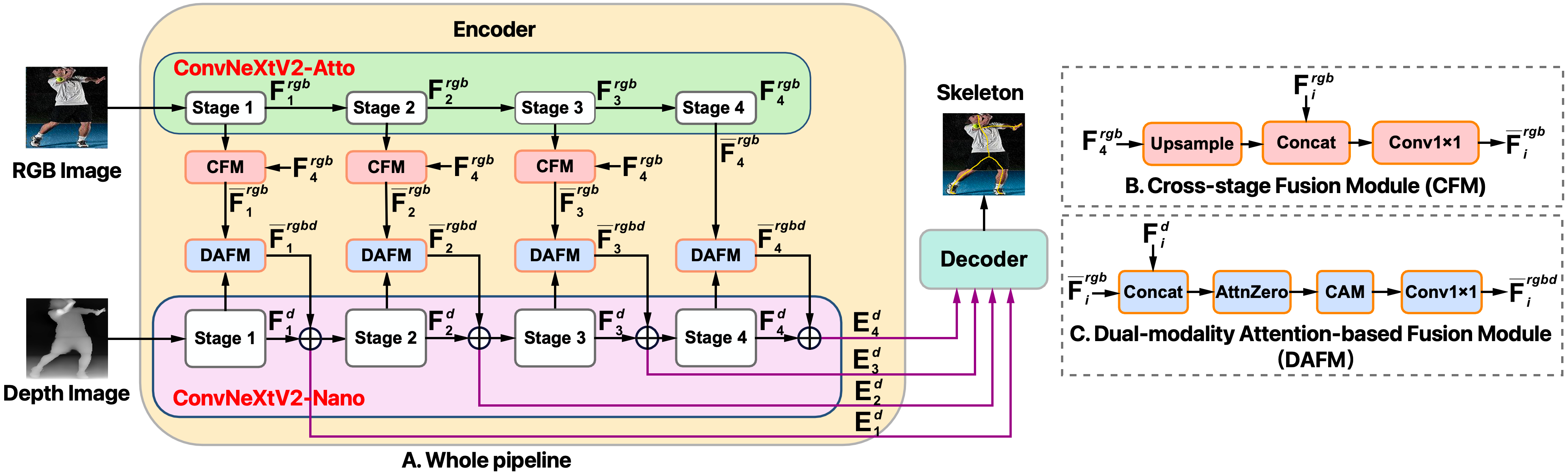}
	\caption{An overview of the DDSkel architecture in (A), which follows an
encoder-decoder structure. The encoder pairs a ConvNeXtV2-Atto backbone for
the RGB image with a ConvNeXtV2-Nano backbone for the depth image. The
CFM first combines the RGB features across
stages, and the DAFM then
uses the RGB cues to enhance the depth features. The resulting depth features are passed to the decoder to
predict skeleton. The detailed structures of CFM and DAFM are shown in (B)
and (C), respectively.
}
	\label{fig:FIG2}
\end{figure*}

\subsection{CNN-based Skeleton Detection} The vast majority of existing deep learning methods for skeleton detection are built upon convolutional neural networks (CNN), centered on manually designed network architectures to boost detection performance. Specifically, Multi-scale Bidirectional Fully Convolutional Network (MSB-FCN) is proposed to better capture and integrate high-level context across multiple scales \cite{Yang03, Yang01}. A hierarchical feature integration mechanism (Hi-Fi) is introduced to refine multi-level feature aggregation \cite{Zhao01}. Fully convolutional networks equipped with scale-associated side outputs are developed to tackle modeling challenges in skeleton prediction \cite{Shen01, Shen02}. DeepFlux predicts two-dimensional vector fields via CNN training, mapping each scene pixel to candidate skeleton points for detection \cite{Wang01, Xu01}. Linear Span Network (LSN) with dedicated linear span units is designed to minimize reconstruction error during skeleton detection \cite{Liu02}. ProMask presents a skeleton probability representation that explicitly encodes skeleton pixels through progressive signals \cite{Bai02, Bai03}. Side-output Residual Network (SRN) leverages error propagation across scales to simplify the fitting of complex outputs with limited network layers \cite{Ke01}, and Rich Side-output Residual Network (RSRN) further fuses side outputs in a deep-to-shallow hierarchical manner to reduce prediction residuals \cite{Liu04}. Beyond manual architecture design, neural architecture search (NAS) has been applied to automatically discover optimal network structures for skeleton extraction \cite{Qiao01, Liu01}. Additional distinctive network designs have also been explored to advance skeleton detection performance \cite{Liu03, Fan01, Fu01, Hou01, Xu02}. 

\subsection{Transformer-based Skeleton Detection} Departing from the dominant CNN paradigm, transformer architectures have recently been introduced to skeleton detection. Unlike most methods that derive results from predicted skeleton heatmaps, BlumNet directly predicts skeletons via a graph decomposition and reconstruction strategy, with an efficient encoder-decoder scheme optimized under an extended transformer architecture \cite{Zhang01}. Additionally, ProFocus revisits the core optimization objective of the skeleton detection task: it proposes to guide the model to gradually shift its focus from coordinate localization regression to point classification during training, and this training paradigm shift can be achieved solely by changing the number of object queries in the Transformer architecture \cite{Wang03}. 

All the aforementioned methods follow the RGB-representation-based feature encoding paradigm and suffer from significant performance degradation when processing complex natural scenes. In contrast, the proposed DDSkel adopts our novel depth-representation-dominant feature encoding paradigm and outperforms all the above methods for skeleton detection on all four mainstream benchmarks, especially in complex scenes.

\section{Methodology}
\subsection{Overview}

Fig.~\ref{fig:FIG2} shows the overall architecture of DDSkel, an
encoder-decoder network whose encoder unidirectionally fuses RGB features
into depth features through an asymmetric dual-branch design. The two
modalities are processed separately by a ConvNeXtV2-Nano depth backbone
 \cite{Woo01} and a ConvNeXtV2-Atto RGB backbone~\cite{Woo01}; the
auxiliary RGB branch has only 23\% as many parameters as the depth branch,
and its four-stage multi-scale features are injected into the corresponding
depth stages through dedicated fusion modules. The augmented depth features
are then fed into an FPN-style decoder that fuses them across scales to
output the skeleton prediction map. 

\subsection{Depth Images Generation}
Depth images for DDSkel are generated by feeding the input RGB images into
the monocular depth estimation model DepthAnythingV2-Small.
For all training and test images, the depth images are generated once and
reused across all stages, so that depth estimation is not repeated during
training or inference.

DDSkel is an open, source-agnostic architecture: the depth estimator is only a replaceable implementation choice, and other monocular depth
estimation methods can be integrated without modifying the network. The architecture can also operate on sensor-acquired depth, subject to standard device-specific preprocessing.

\subsection{The Backbones for Depth and RGB Image}

Both branches adopt the ConvNeXtV2 family~\cite{Woo01}. As a fully
convolutional, hierarchical design, ConvNeXtV2 preserves the two-dimensional
image structure and object boundaries through multi-stage features, and its
fully convolutional masked autoencoder pre-training yields representations
that transfer well to dense prediction tasks, as demonstrated on detection
and segmentation benchmarks. The family also spans a wide range of capacities,
which allows us to build an asymmetric encoder from two architecturally
homogeneous variants.

Specifically, the depth branch uses ConvNeXtV2-Nano (14.98M) as the dominant
encoder, while the auxiliary RGB branch uses ConvNeXtV2-Atto (3.39M), which
contains only 23\% as many parameters as the depth branch yet still provides
semantic cues needed to skeleton detection. Their shared architectural lineage makes the multi-stage features
naturally aligned and easy to fuse, with negligible overhead from the RGB
branch.

Formally, the input RGB image $I_{rgb}$ and the depth image $I_d$, generated
from $I_{rgb}$ by DepthAnythingV2-Small, are passed through their respective
backbones, producing two four-stage feature sets
$\mathbf{F}_{rgb}=\{\mathbf{F}^{rgb}_i\}_{i=1}^{4}$ and
$\mathbf{F}_{d}=\{\mathbf{F}^{d}_i\}_{i=1}^{4}$.

\subsection{Enhancing Depth Features With RGB Features}
Although depth images are inherently better suited to skeleton extraction,
RGB features are still required as auxiliary cues. Since existing skeleton
datasets only annotate specific object categories, the RGB pathway provides
the semantic cues needed to localize the target objects, and it
also offers complementary evidence where the depth image degrades. The 
modules below inject these cues into the depth features.

\subsubsection{Cross-stage Fusion Module}

The Cross-stage Fusion Module (CFM) enriches the RGB features before they
are injected into the depth branch. As the deepest feature $\textbf{F}^{rgb}_4$
carries the strongest high-level semantics, it is upsampled and fused with
each earlier stage $\textbf{F}^{rgb}_1$,$\textbf{F}^{rgb}_2$, and $\textbf{F}^{rgb}_3$, while $\bar{\textbf{F}}^{rgb}_{4}=\textbf{F}^{rgb}_4$ is kept unchanged. For
$i\in\{1,2,3\}$, the CFM is defined as:
\begin{equation}
\bar{\textbf{F}}^{rgb}_{i} = \operatorname{Conv}_{1\times1}\!\big(\operatorname{Cat}(\textbf{F}^{rgb}_i,\operatorname{Up}(\textbf{F}^{rgb}_4))\big),
\label{eq:1}
\end{equation}
where $\operatorname{Up}$ resizes $\textbf{F}^{rgb}_4$ to the resolution of
$\textbf{F}^{rgb}_i$, $\operatorname{Cat}$ concatenates along the channel dimension,
and the $1\times1$ convolution reduces the channels back to that of
$\textbf{F}^{rgb}_i$. This design lets the high-level semantics of $\textbf{F}^{rgb}_4$ guide
every stage, while the multi-stage RGB features provide scale-specific cues
for the depth branch.

\begin{figure*}[t]
  \centering
   \includegraphics[width=1\linewidth]{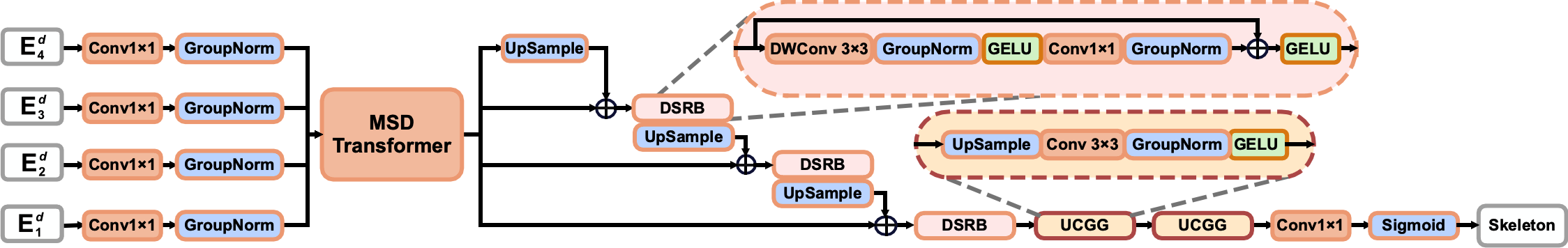}
   \caption{The detailed network structure of the decoder. $\textbf{E}^{d}_1$, $\textbf{E}^{d}_2$, $\textbf{E}^{d}_3$, and $\textbf{E}^{d}_4$ are the output features of the encoder.}
   \label{fig:FIG3}
\end{figure*}

\subsubsection{Dual-modality Attention-based Fusion Module}
The Dual-modality Attention-based Fusion Module (DAFM) is applied at all
four stages. At stage $i$, the feature $\bar{\textbf{F}}^{rgb}_{i}$ is concatenated
with the depth feature $\textbf{F}^{d}_i$ into $\textbf{F}^{rgbd}_i$, which is projected into
query ($Q$), key ($K$), and value ($V$) for spatial attention. We adopt the
linear AttnZero operator~\cite{Li-eccv24}, which provides global spatial
attention at linear computational complexity:
\begin{equation}
\operatorname{AttnZero}
= Q\odot\varphi_2\big(\varphi_2(Q)\,\varphi_2(\beta_1(K)^T\varphi_1(V))\big),
\label{eq:2}
\end{equation}
where $\beta_1$ is the shifted ELU, $\varphi_1$ denotes min-max
normalization, $\varphi_2$ denotes $L_2$ normalization, $\odot$ denotes
element-wise (Hadamard) multiplication, and juxtaposition denotes matrix
multiplication. A channel attention module (CAM) from
CBAM~\cite{woo2018cbam} is then applied, and a $1\times1$ convolution
restores the depth-branch channels, giving
\begin{equation}
\bar{\textbf{F}}^{rgbd}_{i} = \operatorname{Conv}_{1\times1}\big(\operatorname{CAM}(\operatorname{AttnZero}(\textbf{F}^{rgbd}_i))\big).
\label{eq:3}
\end{equation}
The fused feature is added to the depth feature through a residual
connection, yielding the enhanced depth feature
$\textbf{E}^{d}_i=\textbf{F}^{d}_i+\bar{\textbf{F}}^{rgbd}_{i}$, which is sent to the decoder. The dual
spatial-channel attention adaptively weights the RGB and depth cues,
through which the RGB pathway injects category-aware semantic cues and
supplies complementary evidence when the estimated depth locally degrades.
We reuse established, lightweight operators rather than introduce
specialized components.

Furthermore, DAFM is open to the choice of
attention sub-modules: AttnZero and CAM are only the instantiations adopted
in this work, and any other spatial or channel attention module can be
substituted in their place without changing the overall design.

\subsection{Decoder}
As shown in the Fig.~\ref{fig:FIG3}, the decoder consists of a Multi-Scale Deformable Transformer (MSDTransformer) \cite{Zhu-ddetr} and an
FPN-style \cite{Lin-fpn} top-down fusion structure. The four features $\textbf{E}^{d}_1$, $\textbf{E}^{d}_2$, $\textbf{E}^{d}_3$, and $\textbf{E}^{d}_4$ are first
projected and normalized to a common channel dimension through $1\times1$
convolutions and GroupNorm, and the MSDTransformer then aggregates
cross-scale information while preserving their spatial resolutions.
Starting from the coarsest level, the features are progressively upsampled
and added element-wise to the corresponding lateral features, followed by
depthwise separable residual block (DSRB). Each DSRB applies a $3\times3$ depthwise convolution,
GroupNorm, GELU, a $1\times1$ convolution, and GroupNorm in sequence, then
adds the result to the input through a residual connection and passes it
through GELU. Finally, two UCGG blocks (upsampling, $3\times3$
convolution, GroupNorm, GELU) restore the input resolution, and a
$1\times1$ convolution followed by sigmoid outputs the single-channel
skeleton probability map. This design couples the MSDTransformer's
cross-scale shape and semantic context with high-resolution lateral
refinement, enabling accurate pixel-level localization of thin,
topologically continuous skeletons while retaining the low computational
cost of depthwise separable convolutions.

\subsection{Loss Function} 
DDSkel is optimized with a weighted binary cross-entropy
loss~\cite{xie2015hed} that mitigates the severe skeleton/background pixel
imbalance, together with a soft Dice loss~\cite{Milletari} that enforces
region-level overlap. 

\section{Experiments}
\subsection{Datasets and Evaluation Protocol}
We conduct experiments on skeleton detection using four widely-used datasets: \textbf{SK-LARGE} \cite{Shen01}, \textbf{SK506} \cite{Shen02}, \textbf{WH-SYMMAX} \cite{Shen03}, and \textbf{SymPASCAL} \cite{Ke01}. Among them, the SymPASCAL contains a large volume of complex uncropped in-the-wild images with cluttered backgrounds and dense distracting textures, making it more challenging and difficult than other three datasets. The datasets splits are as follows: SK506 contains 300 training images and 206 testing images; WH-SYMMAX contains 328 images, where the first 228 images are used for training and the remaining 100 images are used for testing; SK-LARGE contains 746 training images and 745 testing images; SymPASCAL contains 648 training images and 787 testing images.

Like other state-of-the-art methods, in our experiments, we use the F-measure metric \cite{Wang01} to evaluate the performance of skeleton detection.
\subsection{Implementation Details}

\subsubsection{Details of Training}
During training, both RGB and depth backbones are initialized with official pretrained weights and jointly optimized with the rest of the DDSkel. The effective batch size is 8 (batch size 2, four accumulation steps). We adopt AdamW with module-specific parameter groups. The learning rate is set to $1\times10^{-4}$ for both pretrained backbones and $3\times10^{-4}$ for other parts of the DDSkel. The weight decay is $1\times10^{-2}$. The model is trained for 80 epochs. Training randomly selects the image long side from $\{320,384,448\}$ while preserving the aspect ratio. RGB, depth, and ground truth images share horizontal flips and affine augmentation. RGB and depth images are normalized using ImageNet and training-set statistics, respectively. 

\subsubsection{Details of Testing}

Testing uses a single forward pass with a long-side resolution of 384. Cropped logits are bilinearly resized to the original resolution and passed through sigmoid. Following prior methods, we apply directional NMS, thresholding, and thinning at the original resolution.

\subsubsection{Details of Encoder and Decoder}

The depth-primary encoder is ConvNeXtV2-Nano, which contains four stages with channel dimensions $[80,160,320,640]$ and block depths $[2,2,8,2]$. The RGB auxiliary encoder is ConvNeXtV2-Atto, with channel dimensions $[40,80,160,320]$ and block depths $[2,2,6,2]$. The single-channel depth image is replicated across three channels to reuse the pretrained input stem. MSDTransformer and the FPN use 96 channels, with $1\times1$ input projections followed by GroupNorm (32 groups). MSDTransformer has 2 layers, 4 heads, and 4 sampling points per head per level, with LayerNorm and a $96\to192\to96$ FFN using ReLU. The FPN contains 3 DSRBs with $3\times3$ depthwise and $1\times1$ pointwise kernels. DSRBs and normalized head convolutions use GroupNorm (8 groups) and GELU. The head uses two $2\times$ bilinear upsampling stages with channels $96\to48\to24\to1$,  and a final $1\times1$ projection.

\subsubsection{Details of Evaluation}

We use 49 thresholds uniformly spaced from 0.02 to 0.98. At each threshold, the NMS-processed probability map is binarized and refined using Guo--Hall thinning. Dataset-level precision and recall are computed from one-to-one skeleton-pixel matches within $1\%$ of the original image diagonal. We report the dataset-level optimal F-measure (ODS), together with its corresponding precision and recall. All comparisons and ablations use the same evaluation protocol.

\subsection{Comparison with State-of-the-Art Methods}
\subsubsection{Quantitative Comparison}
Table~\ref{tab:ddskel:1} presents a quantitative comparison between our
DDSkel and eight state-of-the-art methods
(HED~\cite{xie2015hed}, SRN~\cite{Ke01}, Hi-Fi~\cite{Zhao01},
DeepFlux~\cite{Wang01}, FMRN~\cite{Fan01}, ProMask~\cite{Bai02},
BlumNet~\cite{Zhang01}, and ProFocus~\cite{Wang03}) for skeleton detection
on four mainstream datasets under the original ground-truth annotations,
evaluated by the ODS F-measure. Values highlighted in bold correspond to
the highest-ranked F-measure scores. As can be observed from the table, on
SymPASCAL, which contains a large number of challenging images, DDSkel
achieves significantly superior performance over all competing methods.
Its F-measure reaches 0.644 and outperforms the second-best method,
ProMask (0.582), by 0.062 in absolute terms (about 10.7\% relatively),
yet DDSkel only uses 19.25M trainable parameters, roughly about 19\% of
BlumNet (99.32M) and ProFocus (100.84M), as shown in Table~\ref{tab:ddskel:2}. This verifies
that DDSkel yields notably better performance with far fewer parameters on
complex images.

Furthermore, on the remaining three datasets with much fewer complex images, DDSkel still outperforms all competing methods. Specifically, on WH-SYMMAX, DDSkel leads the second-best BlumNet by 0.021 (about 2.4\%), while on SK506 and SK-LARGE, DDSkel outperforms BlumNet by 0.006 each (relative gains of about 0.8\% and 0.7\%, respectively). Across all four datasets, DDSkel consistently outperforms the methods with a comparable parameter count (see Table~\ref{tab:ddskel:2}).

\begin{table}[t]
\caption{Quantitative comparison (ODS F-measure) on four mainstream datasets against eight state-of-the-art methods, using the original ground-truth annotations. Bold denotes the best result.}
\label{tab:ddskel:1}
\begingroup
\centering
\footnotesize
\setlength{\tabcolsep}{3pt}
\renewcommand{\arraystretch}{1.05}
\begin{tabular*}{\linewidth}{@{\extracolsep{\fill}}lcccc@{}}
\toprule
\textbf{Model} & \textbf{SK506} & \textbf{SK-LARGE} & \textbf{WH-SYMMAX} & \textbf{SymPASCAL} \\
\midrule
HED \cite{xie2015hed} & 0.542 & 0.497 & 0.732 & 0.369 \\
SRN \cite{Ke01} & 0.632 & 0.658 & 0.780 & 0.443 \\
Hi-Fi \cite{Zhao01} & 0.681 & 0.724 & 0.805 & 0.454 \\
DeepFlux \cite{Wang01} & 0.695 & 0.732 & 0.840 & 0.502 \\
FMRN \cite{Fan01} & 0.751 & 0.789 & 0.865 & 0.508 \\
ProMask \cite{Bai02} & 0.754 & 0.772 & 0.875 & 0.582 \\
BlumNet \cite{Zhang01} & 0.773 & 0.813 & 0.888 & 0.559 \\
ProFocus \cite{Wang03} & 0.755 & 0.802 & 0.885 & 0.576 \\
\textbf{DDSkel} & \textbf{0.779} & \textbf{0.819} & \textbf{0.909} & \textbf{0.644} \\
\bottomrule
\end{tabular*}
\endgroup
\end{table}

\begin{table}[t]
\caption{Comparison of model complexity and inference cost. Total
includes frozen parameters, while Trainable counts only trainable ones;
w/o DAV2 excludes the frozen DepthAnythingV2-Small and w/ DAV2 includes it.}
\label{tab:ddskel:2}
\begingroup
\centering
\footnotesize
\setlength{\tabcolsep}{3pt}
\renewcommand{\arraystretch}{1.05}
\begin{tabular*}{\linewidth}{@{\extracolsep{\fill}}lcccc@{}}
\toprule
\textbf{Model} & \textbf{Total(M)} & \textbf{Trainable(M)} & \textbf{FLOPs(G)} & \textbf{Time(ms)} \\
\midrule
HED \cite{xie2015hed} & 14.72 & 14.72 & 103.75 & 5.18 \\
SRN \cite{Ke01} & 14.72 & 14.72 & 103.75 & 5.23 \\
DeepFlux \cite{Wang01} & 18.32 & 18.32 & 125.93 & 6.39 \\
BlumNet \cite{Zhang01} & 99.32 & 99.32 & 254.45 & 23.03 \\
ProFocus \cite{Wang03} & 100.84 & 100.84 & 391.43 & 44.12 \\
\textbf{DDSkel (w/o DAV2)} & \textbf{19.25} & \textbf{19.25} & \textbf{28.92} & \textbf{4.01} \\
\textbf{DDSkel (w/ DAV2)} & \textbf{44.04} & \textbf{19.25} & \textbf{144.18} & \textbf{10.96} \\
\bottomrule
\end{tabular*}
\endgroup
\end{table}

Table~\ref{tab:ddskel:2} compares model complexity and inference cost
under two accounting scopes. When the depth generator is excluded
(w/o DAV2), DDSkel requires only 19.25M parameters, 28.92G FLOPs, and
4.01ms, i.e. merely 7.4\% of the FLOPs and 9.1\% of the latency of
ProFocus. Even when the frozen DepthAnythingV2-Small is included for an
end-to-end view (w/ DAV2), DDSkel incurs 44.04M total parameters,
144.18G FLOPs, and 10.96ms, still markedly cheaper than both BlumNet
(254.45G, 23.03ms) and ProFocus (391.43G, 44.12ms). Combining
Table~\ref{tab:ddskel:1} and Table~\ref{tab:ddskel:2}, DDSkel attains the
best detection performance among methods with comparable or smaller
parameter scales.

Additionally, Yang et al.~\cite{Yang02} point out that some skeleton
annotations of these four datasets suffer from certain quality issues.
To address this, they re-annotate all four datasets. To verify whether
our conclusion still holds under the refined annotations, we further
repeat the quantitative comparison in Table~\ref{tab:ddskel:1} on the
re-annotated versions of all four datasets, retraining and re-evaluating
all competing methods under the same protocol. On these re-annotated
benchmarks, DDSkel again achieves the highest F-measure on every dataset
and consistently outperforms the compared methods,
indicating that its advantage is not tied to the original, potentially
noisy annotations.

\subsubsection{Qualitative Comparison}

\begin{figure}[t]
	\centering
	\includegraphics[width=1\columnwidth]{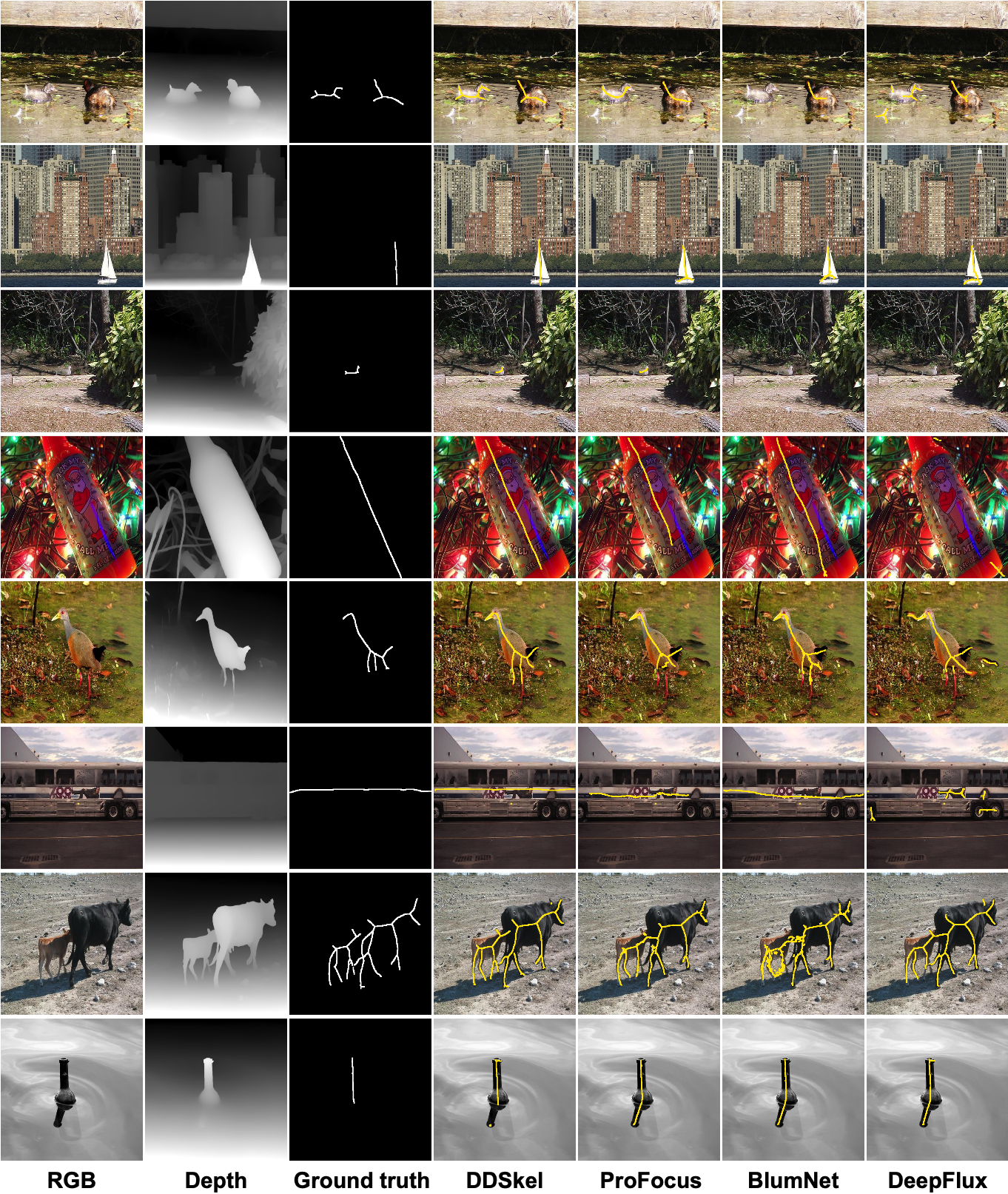}
	\caption{The qualitative comparison between DDSkel and three state-of-the-art methods, ProFocus \cite{Wang03}, BlumNet \cite{Zhang01}, and DeepFlux \cite{Wang01}.
	}
	\label{fig:FIG4}
\end{figure}

Fig.~\ref{fig:FIG4} compares DDSkel with three state-of-the-art methods
(ProFocus~\cite{Wang03}, BlumNet~\cite{Zhang01}, and
DeepFlux~\cite{Wang01}) on challenging scenes. Although the RGB images
contain cluttered textures, occlusions, and foreground-background
confusion, these interferences are largely suppressed in the depth images,
where the target contours remain clear. As a result, the skeletons produced
by DDSkel best match the ground truth, with fewer broken branches and
spurious responses than the competing methods, which exhibit visible
misalignment and extra or missing parts. 

\subsection{Ablation Study}
\begin{table}[b]
\caption{Ablation of the dual-backbone design and depth primacy. For dual-branch settings, the first modality is assigned to the dominant backbone and the second to the auxiliary branch.}
\label{tab:ddskel:3}
\begingroup
\centering
\footnotesize
\setlength{\tabcolsep}{3pt}
\renewcommand{\arraystretch}{1.05}
\begin{tabular*}{\linewidth}{@{\extracolsep{\fill}}lcc@{}}
\toprule
\textbf{Configuration} & \textbf{SK-LARGE} & \textbf{SymPASCAL} \\
\midrule
RGB-Only & 0.778 & 0.517 \\
Depth-Only & 0.807 & 0.614 \\
RGB--RGB & 0.790 & 0.552 \\
Depth--Depth & 0.813 & 0.633 \\
RGB--Depth & 0.816 & 0.627 \\
\textbf{Depth--RGB(DDSkel)} & \textbf{0.819} & \textbf{0.644} \\
\bottomrule
\end{tabular*}
\endgroup
\end{table}

\subsubsection{Ablation on Dual-backbone and Depth Primacy} 
This ablation study justifies the dual-backbone design of the encoder
with both RGB and depth modalities, as well as the rationale for adopting
depth as the dominant modality. Table~\ref{tab:ddskel:3} presents the
results under a fixed architecture and training protocol.

In the first column of Table~\ref{tab:ddskel:3}, each variant is defined as follows.
RGB-Only: only the larger backbone takes RGB images, with the smaller
backbone removed. Depth-Only: only the larger backbone takes depth images,
with the smaller backbone removed. RGB-RGB: both backbones are fed with RGB
images. Depth-Depth: both backbones are fed with depth images. RGB-Depth:
the larger backbone takes RGB while the smaller takes depth. The last two
columns report the F-measure on SK-LARGE and SymPASCAL.

Among the single-backbone baselines, Depth-Only clearly beats RGB-Only on
SymPASCAL (0.614 vs 0.517, +0.097), indicating that depth is a more
suitable primary representation for skeleton extraction. The gain is not
due to capacity: duplicating the same modality (Depth-Depth) improves
only marginally over Depth-Only (0.633 vs 0.614), whereas, at the
identical dual-backbone budget, replacing the auxiliary depth with an
RGB branch raises the score further (Depth-RGB 0.644 vs Depth-Depth
0.633), confirming complementary rather than duplicated information.
Depth-RGB is consistently preferable to the role-reversed RGB-Depth on
both datasets (0.644 vs 0.627 on SymPASCAL), favoring the depth-primary
assignment while RGB serves as the auxiliary. The same ordering holds on SK-LARGE, where all other dual-branch configurations remain below the depth-dominant design, ruling out branch
capacity and fusion multiplicity as the cause.

\subsubsection{Ablation on RGB Auxiliary Backbone} 
In this ablation, we replace the RGB auxiliary backbone with three larger
ConvNeXtV2 variants (Nano, Pico, and Femto) while keeping the depth
backbone and fusion configuration fixed, so only the RGB-branch capacity
is varied. Table~\ref{tab:table4} shows that enlarging the RGB branch yields no
consistent gain: the F-measure fluctuates across variants and improves
only marginally, while the parameter count rises sharply. In particular,
relative to Nano, Atto reduces the RGB encoder from 14.98M to 3.39M
(a 77.4\% reduction) while lowering the F-measure by only 0.002 on
SK-LARGE and 0.001 on SymPASCAL. Enlarging the auxiliary branch is
therefore an unfavorable trade-off, and we adopt Atto as the default for
its favorable balance between performance and efficiency.

\begin{table}[t]
\caption{Ablation of RGB auxiliary-branch capacity with the depth backbone and fusion configuration fixed. Parentheses denote RGB encoder parameters, excluding the classification head.}
\label{tab:table4}
\begingroup
\centering
\footnotesize
\setlength{\tabcolsep}{2.5pt}
\renewcommand{\arraystretch}{1.05}
\begin{tabular*}{\linewidth}{@{\extracolsep{\fill}}lcc@{}}
\toprule
\textbf{RGB Auxiliary Branch} & \textbf{SK-LARGE} & \textbf{SymPASCAL} \\
\midrule
ConvNeXtV2-Nano (14.98M) & \textbf{0.821} & \textbf{0.645} \\
ConvNeXtV2-Pico (8.55M) & 0.812 & 0.641 \\
ConvNeXtV2-Femto (4.85M) & 0.817 & 0.640 \\
\textbf{ConvNeXtV2-Atto (3.39M, Ours)} & 0.819 & 0.644 \\
\bottomrule
\end{tabular*}
\endgroup
\end{table}

\subsubsection{Ablation on Depth Estimation Models} 
In this ablation, we replace the depth estimator with four alternatives (MoGe-2-Small \cite{WangMoge2}, YOLO26s-Depth \cite{JocherYOLO26}, ZipDepth \cite{TosiZipDepth}, and DepthART \cite{XueDepthART}) while keeping the DDSkel architecture and fusion configuration fixed. Table~\ref{tab:table5} shows that DDSkel remains competitive across all estimators, indicating that the architecture accommodates depth cues from different sources rather than depending on a single generator. The F-measure still varies across estimators, with a wider span on SymPASCAL (0.060) than on SK-LARGE (0.024), and no strictly monotonic trend is observed, suggesting that complex scenes are more sensitive to the depth representation. DepthAnythingV2-Small yields the highest score on both datasets (0.819 and 0.644), and is therefore adopted as the default.

\begin{table}[t]
\caption{Ablation of the depth estimation model used to provide geometric cues for DDSkel. Parentheses denote the parameter count of each depth estimator.}
\label{tab:table5}
\begingroup
\centering
\footnotesize
\setlength{\tabcolsep}{2pt}
\renewcommand{\arraystretch}{1.05}
\begin{tabular*}{\linewidth}{@{\extracolsep{\fill}}lcc@{}}
\toprule
\textbf{Depth model} & \textbf{SK-LARGE} & \textbf{SymPASCAL} \\
\midrule
MoGe-2-Small (35.10M) & 0.812 & 0.627 \\
YOLO26s-Depth (13.21M) & 0.795 & 0.591 \\
ZipDepth (6.14M) & 0.795 & 0.584 \\
DepthART (6.05M) & 0.799 & 0.600 \\
\textbf{DepthAnythingV2-Small (24.79M, Our)} & \textbf{0.819} & \textbf{0.644} \\
\bottomrule
\end{tabular*}
\endgroup
\end{table}

\subsubsection{Ablation on CFM and DAFM}

In this ablation, we remove CFM, DAFM, or both to isolate their
contributions, while keeping the rest of DDSkel unchanged. As shown in
Table~\ref{tab:table6}, retaining either module alone improves over the baseline on both
datasets, confirming that CFM and DAFM are both beneficial. The complete model with both
modules performs best, and the gain is larger on SymPASCAL (0.644 vs
0.621, +0.023) than on SK-LARGE (+0.008), indicating that the two modules,
which handle cross-stage semantic enrichment and cross-modal injection
respectively, provide complementary rather than redundant benefits. We
therefore retain both modules in the default design.

\begin{table}[t]
\caption{Effects of cross-stage RGB fusion and dual-modality attention in DDSkel, assessed by removing CFM, DAFM, or both.}
\label{tab:table6}
\begingroup
\centering
\footnotesize
\setlength{\tabcolsep}{3pt}
\renewcommand{\arraystretch}{1.05}
\begin{tabular*}{\linewidth}{@{\extracolsep{\fill}}lcc@{}}
\toprule
\textbf{Configuration} & \textbf{SK-LARGE} & \textbf{SymPASCAL} \\
\midrule
Without CFM and DAFM & 0.811 & 0.621 \\
Without DAFM & 0.814 & 0.631 \\
Without CFM & 0.815 & 0.628 \\
\textbf{DDSkel} & \textbf{0.819} & \textbf{0.644} \\
\bottomrule
\end{tabular*}
\endgroup
\end{table}

\subsection{Robustness Analysis of Depth Degradation}

\begin{figure}[t]
	\centering
	\includegraphics[width=0.8\columnwidth]{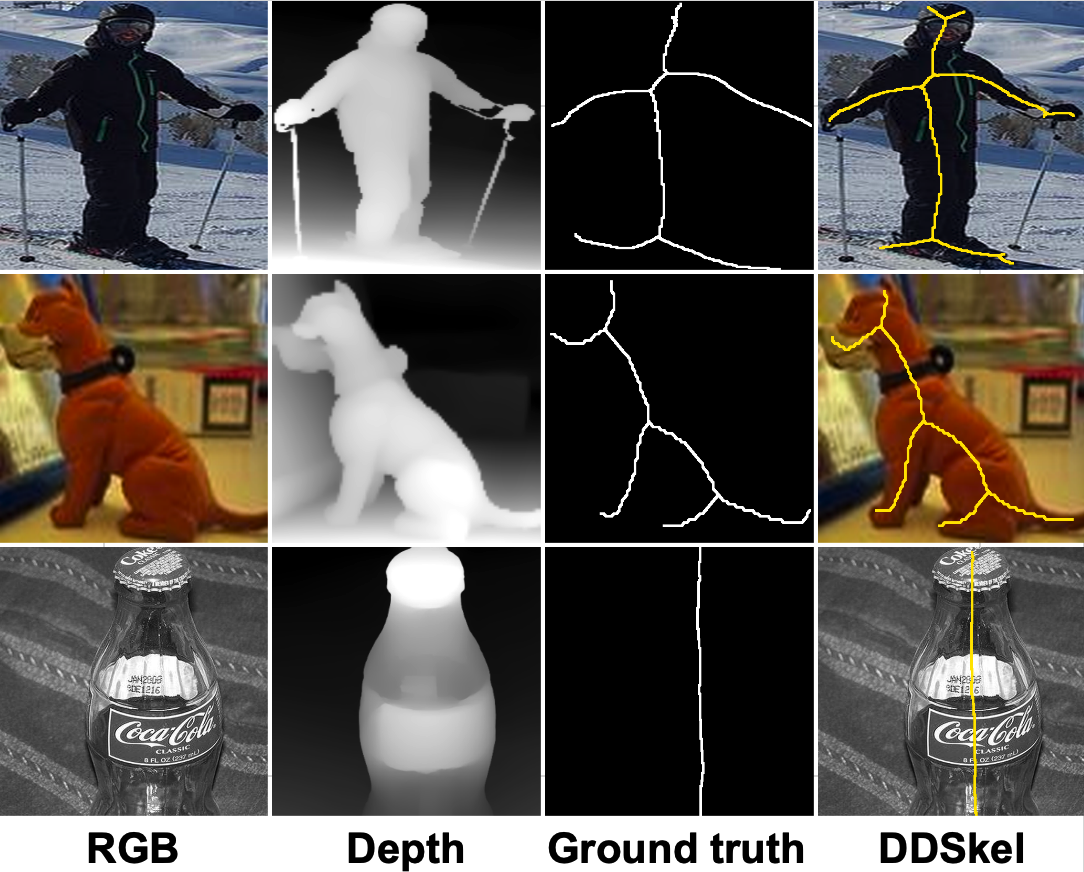}
	\caption{Qualitative results of DDSkel on depth images containing typical
artifacts.
	}
	\label{fig:FIG5}
\end{figure}

DepthAnythingV2-Small generally produces high-quality depth images across the four
mainstream skeleton detection datasets, yet a few depth images still contain
errors. Fig.~\ref{fig:FIG5} shows how DDSkel behaves on such cases.
In the first row, the white portion on the sleeve of the black jacket is
similar in appearance to the snowy background, inducing background-like
responses around the arm; nevertheless, the predicted arm skeleton remains
largely continuous. In the second row, a depth shift occurs at the color
boundary between the toy dog's muzzle and head, but the main skeletal
connections among the muzzle, head, and torso are preserved. In the third
row, the transparent bottle shows irregular depth partitions due to its
material, while the model still extracts a continuous longitudinal axis.
These examples suggest that DDSkel tolerates common depth-estimation
artifacts and retains the main skeletal structure under imperfect geometric
priors, which benefits from the depth-primary, RGB-auxiliary design of
DDSkel.

\subsection{Failure Case Analysis}

\begin{figure}[t]
	\centering
	\includegraphics[width=1\columnwidth]{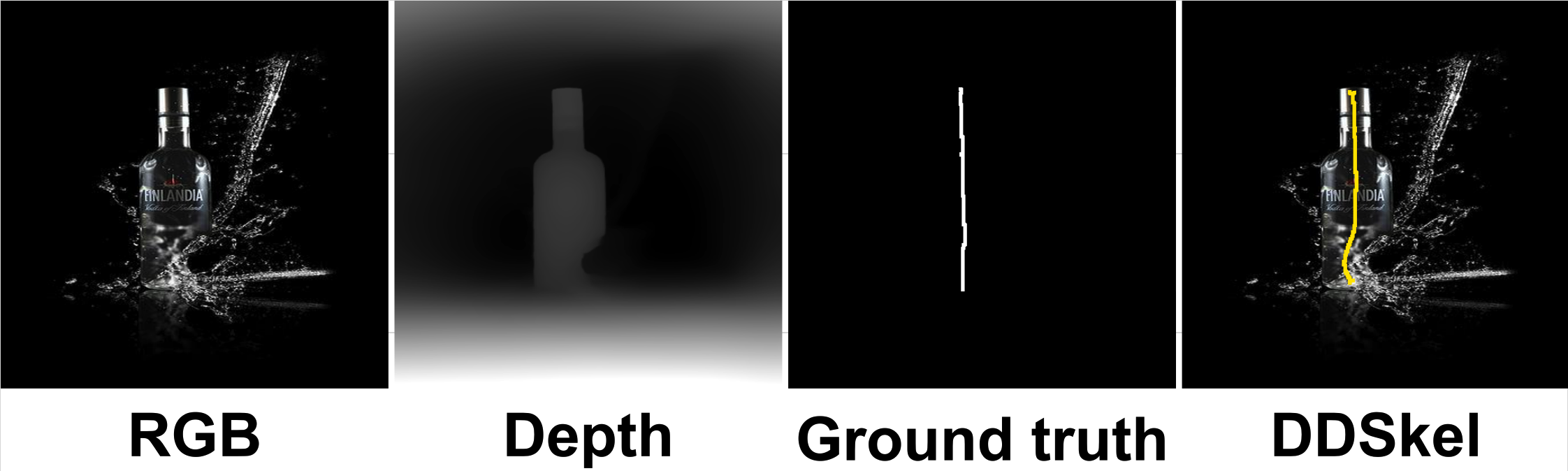}
	\caption{A failure case where both modalities provide ambiguous cues.
	}
	\label{fig:FIG6}
\end{figure}

Fig.~\ref{fig:FIG6} presents a challenging example in which both
modalities provide ambiguous cues. Reflections and water splashes interfere
with the lower part of the transparent bottle, making its surface boundaries
difficult to distinguish in RGB. Meanwhile, the estimated depth image shows
an irregular contraction of the bottle silhouette on its lower-right side,
which further complicates skeleton detection. Consequently, the lower part of
the predicted skeleton bends leftward and deviates from the approximately
vertical target axis. This indicates that when RGB cannot supply sufficiently
clear complementary cues and the depth prior also contains shape distortions,
the ability of DDSkel to recover the correct skeleton is limited. Such cases
are mainly associated with transparent or reflective materials, and could be
addressed by stronger depth estimation or material-aware modeling in future
work.

\section{Conclusion}
This paper presents DDSkel, a natural-scene skeleton detector built on a
novel depth-representation-dominant feature encoding paradigm that departs
from the long-standing RGB-representation-based paradigm. By coupling a
dominant depth pathway with a lightweight auxiliary RGB pathway, DDSkel
attains the highest F-measure on all four mainstream benchmarks with only about
19\% of the trainable parameters of the largest state-of-the-art methods, demonstrating
that the depth-dominant paradigm is a more suitable and parameter-efficient
choice for skeleton detection. Looking forward, our future work will expand and diversify existing benchmarks, and more fully exploit the potential of depth representations for skeleton detection, including stronger depth estimation and material-aware modeling for the transparent and reflective cases discussed above.

{\small
\bibliographystyle{ieeenat_fullname}
\bibliography{main}
}
\end{document}